\documentclass[conference]{IEEEtran}
\IEEEoverridecommandlockouts
\usepackage{cite}
\usepackage{amsmath,amssymb,amsfonts}
\usepackage{algorithmic}
\usepackage{graphicx}
\usepackage{textcomp}
\usepackage{xcolor}
\def\BibTeX{{\rm B\kern-.05em{\sc i\kern-.025em b}\kern-.08em
    T\kern-.1667em\lower.7ex\hbox{E}\kern-.125emX}}

\usepackage[hidelinks]{hyperref}
\usepackage{stfloats}

\usepackage{subfig}

\usepackage{flushend}

\begin{document}

\title{
	SOMOSPIE: A modular SOil MOisture SPatial Inference Engine based on data driven decisions
	\thanks{
		M.~G. acknowledges support from a Conacyt fellowship; M.~T. and R.~V. acknowledge support from NSF (\#1724843, \#1854312); and R.~V. acknowledges partial support from NSF (\#1652594).
	}
}

\author{\IEEEauthorblockN{Danny Rorabaugh}
	\IEEEauthorblockA{\textit{Electrical Engineering and Computer Science} \\
		\textit{University of Tennessee, Knoxville}\\
		dror@utk.edu}
	\and
	\IEEEauthorblockN{Mario Guevara}
	\IEEEauthorblockA{\textit{Plant and Soil Sciences} \\
		\textit{University of Delaware, Newark}\\
		mguevara@udel.edu}
	\and
	\IEEEauthorblockN{Ricardo Llamas}
	\IEEEauthorblockA{\textit{Plant and Soil Sciences} \\
		\textit{University of Delaware, Newark}\\
		rllamas@udel.edu}
	\and
	\IEEEauthorblockN{Joy Kitson}
	\IEEEauthorblockA{\textit{Computer and Information Sciences} \\
		\textit{University of Delaware, Newark}\\
		tkitson@udel.edu}
	\and
	\IEEEauthorblockN{Rodrigo Vargas}
	\IEEEauthorblockA{\textit{Plant and Soil Sciences} \\
		\textit{University of Delaware, Newark}\\
		rvargas@udel.edu}
	\and
	\IEEEauthorblockN{Michela Taufer}
	\IEEEauthorblockA{\textit{Electrical Engineering and Computer Science} \\
		\textit{University of Tennessee, Knoxville}\\
		taufer@gmail.com}
}

\maketitle

\begin{abstract}
	The current availability of soil moisture data over large areas comes from satellite remote sensing technologies (i.e., radar-based systems), but these data have coarse resolution and often exhibit large spatial information gaps. Where data are too coarse or sparse for a given need (e.g., precision agriculture), one can leverage machine-learning techniques coupled with other sources of environmental information (e.g., topography) to generate gap-free information and at a finer spatial resolution (i.e., increased granularity). To this end, we develop a spatial inference engine consisting of modular stages for processing spatial environmental data, generating predictions with machine-learning techniques, and analyzing these predictions. We demonstrate the functionality of this approach and the effects of data processing choices via multiple prediction maps over a United States ecological region with a highly diverse soil moisture profile (i.e., the Middle Atlantic Coastal Plains). The relevance of our work derives from a pressing need to improve the spatial representation of soil moisture for applications in environmental sciences (e.g., ecological niche modeling, carbon monitoring systems, and other Earth system models) and precision agriculture (e.g., optimizing irrigation practices and other  land management decisions).
\end{abstract}



\section{Introduction}

Soil moisture is a critical variable that links climate dynamics with water and food security. 
It regulates land-atmosphere interactions (e.g., via evapotranspiration--the loss of water from evaporation and plant transpiration to the atmosphere), and it is directly linked with plant productivity and plant survival \cite{Koster2004}. 
Information on soil moisture is important to design appropriate irrigation strategies to increase crop yield, and long-term soil moisture coupled with climate information provides insights into trends and potential agricultural thresholds and risks \cite{Denmead1962, Bauch2016, Phillips2014}. 
Thus, information on soil moisture is needed to assess the implications of environmental variability and consequently is a key factor to inform and enable precision agriculture. 
Currently, large areas of western states of the conterminous United States (CONUS) are experiencing an exceptional drought, and most information on water limitation has been derived from changes in precipitation patterns \cite{Griffen2014}. 
This is just one example of where soil moisture information can contribute to situations of critical importance. 

\begin{figure*}[htbp]
	\centerline{\includegraphics[width=\textwidth]{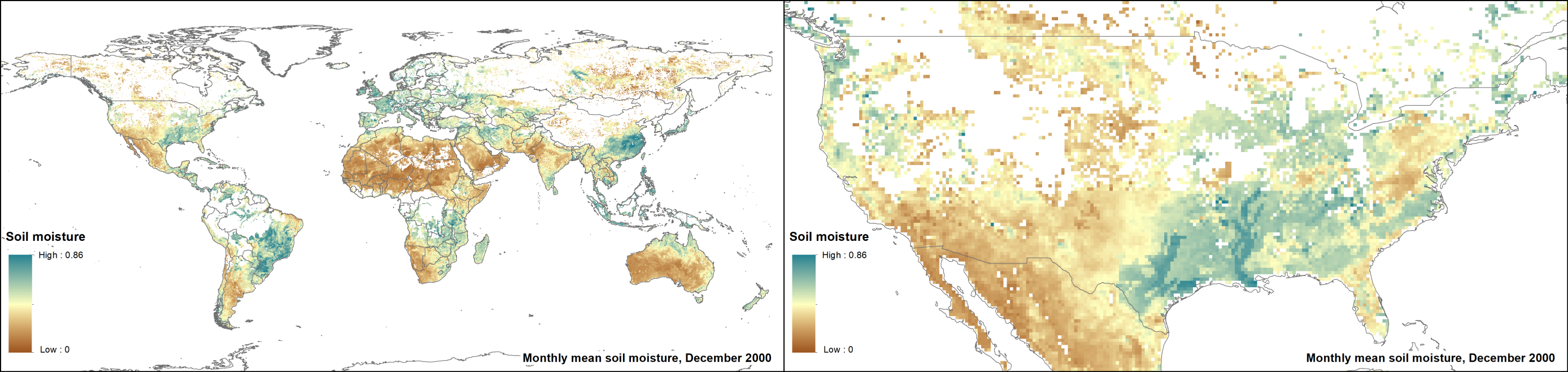}}
	\caption{Monthly soil moisture (m$^3$/m$^3$) averages for December 2000 with gaps where data cannot be collected accurately because of dense vegetation, snow cover, and extremely dry surfaces. Averaged from daily data from the ESA-CCI soil moisture database (\url{esa-soilmoisture-cci.org}).}
	\label{fig-gaps}
\end{figure*}

The current availability in soil moisture data over large areas comes from remote sensing (i.e., satellites with radar sensors), which provides nearly global coverage of soil moisture at spatial resolution of tens of kilometers~\cite{Entekhabi2010, Dorigo2011}.  
Recent efforts are devoted to increase the spatial resolution of current estimates (\url{smap.jpl.nasa.gov}). 
Other efforts have focused on harmonizing historical satellite soil moisture records for larger periods of time and from several information sources (\url{esa-soilmoisture-cci.org}). 
Satellite soil moisture data has two main shortcomings. 
First, although satellites can provide daily global information, they are limited to coarse spatial resolution (at the multi-kilometer scale). 
Second, satellites are unable to measure soil moisture in areas of dense vegetation, snow cover, or extremely dry surfaces; this results in gaps in the data. 
Fig.~\ref{fig-gaps} shows an example of the monthly averages of daily soil moisture data for December 2000. 
The figure shows examples of spatial information gaps across the globe due to, for example, dense vegetation over the Amazon region and central Africa.  
To use the spatial representation of soil moisture for applications in environmental sciences (e.g., ecological niche modeling, carbon monitoring systems, and other Earth system models) and precision agriculture (e.g., optimizing irrigation practices and other land management decisions), we need to increase the spatial resolution of information and predict values in areas with missing data. 

In this paper, we address the two shortcomings associated with satellite data (i.e., coarse-grained resolution and spatial information gaps) by providing a modular SOil MOisture SPatial Inference Engine (SOMOSPIE).  
SOMOSPIE consists of modular components including input of available data at its native spatial resolution, selection of a geographic region of interest, prediction of missing values across the entire region of interest (i.e., gap-filling), analysis of generated predictions, and visualization of both predictions and analyses. 
To predict soil moisture, our engine leverages hydrologically meaningful terrain parameters (e.g., slope and topographic wetness index) calculated using an open source platforms for standard terrain analysis (i.e., SAGA-GIS) 
 and various machine learning methods. 
The engine combines the publicly available datasets of satellite-derived soil moisture measurements from the European Space Agency (ESA) and generates fine-grained, gap-free soil moisture predictions using three implementations of machine learning algorithms: 
a kernel-based approach (kernel-weighted k-nearest neighbors or KKNN), 
the Hybrid Piecewise Polynomial approach (HYPPO), and 
a tree-based approach (Random Forests or RF). 
Data processing functionality of our engine includes selection of a region of interest, which we demonstrate using ecoregions as defined for North America by the Commission for Environmental Cooperation \cite{CEC2009}.  
We exhibit the full functionality of our engine on the Middle Atlantic Coastal Plains in the eastern United States, a region with a diverse soil moisture profile. 

The main contributions of this paper are: 
\begin{enumerate}
    \item A spatial inferences engine (SOMOSPIE) and all the data and components needed to generate viable soil moisture predictions; 
    \item An empirical study of the engine's functionality including an assessment of data processing and fine-grained predictions over a United States ecological region with a highly diverse soil moisture profile (i.e., the Middle Atlantic Coastal Plains).
\end{enumerate}

\begin{table*}[hbp]
	\caption{List of datasets used in this study.}
	\begin{center}
		\begin{tabular}{|c|c|c|c|c|}
			\hline
			\textbf{Dataset} & \textbf{Spatial resolution} 
			& \textbf{Temporal resolution} 
			& \textbf{Variable / Description}
			& \textbf{Source}
			\\ 
			\hline
			ESA-CCI & 0.25 degrees & Daily, 1978--2016 
			& soil moisture (m\textsuperscript{3}/m\textsuperscript{3}) & European Space Agency
			\\ 
			\hline
			DSM & $\approx\!30$ meters & Static (`Current')
			& Land surface characteristics
			& The Japan Aerospace Exploration Agency 
			\\ 
			\hline
			CEC & n/a & Static (`Current') & Ecoregion boundaries
			& Commission for Environmental Cooperation
			\\ 
			\hline
		\end{tabular}
		\label{tab-data}
	\end{center}
\end{table*}

The rest of the paper is structured as follows: 
Section~\ref{sec-data} describes the datasets used for this project. 
Section~\ref{sec-model} consists of a breakdown of the components of SOMOSPIE throughout its three stages: data processing, prediction generation, and prediction analysis and visualization. 
Section~\ref{sec-results} contains results, including soil moisture predictions that leverage various modular elements of SOMOSPIE and analyses thereof. 
Section~\ref{sec-related} brings us around with related work. 
Section~\ref{sec-conclusion} wraps up this paper with our conclusion.

\section{Our Datasets} \label{sec-data}

Our work builds upon publicly available data collections associated with remotely-sensed soil moisture information, topographic characteristics derived from quantitative land surface analysis, and eco-regionalization of North America. 
These diverse datasets are cornerstones of SOMOSPIE: (a) moisture records, (b) a digital surface model (DSM), and (c) boundaries for ecoregions. 
Table~\ref{tab-data} reports the data resolutions and sources. 

\begin{figure}[htbp]
	\centerline{\includegraphics[width=.5\textwidth]{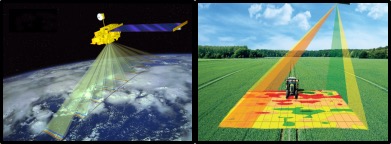}}
	\caption{Illustration of how satellites collect raster data across the surface of the Earth \cite{Kitson2017, Liu2011}.}
	\label{fig-sat}
\end{figure}

Satellite-derived soil moisture data were downloaded from the ESA-CCI soil moisture initiative \cite{Dorigo2011, Wagner2012, Liu2011, Liu2012}. 
The ESA-CCI soil moisture data are collected in a raster format and have an original spatial resolution of $0.25 \times 0.25$ lat-lon degrees (about $27 \times 27$ km). Fig.~\ref{fig-sat} portrays a satellite collecting raster data. 
Each pixel in the raster file corresponds to a square of land and contains the satellite-derived soil moisture value for that land surface. 
The value is a ratio (between 0 and 1), the number of m\textsuperscript{3} of water per m\textsuperscript{3} of surface soil, where 0 indicates dry soil and 1 indicates water-saturated soil. 
This dataset is representative for the first 0 to 5 cm of soil surface \cite{Hirschi2014}. 
The original temporal scale of the ESA-CCI is daily, but for this study we move from daily to monthly time steps by averaging all daily values in a given pixel across an entire month.

Our topography dataset, consisting of multiple terrain parameters, is based on DSMs, which are available at several resolutions and are useful to represent multiple terrain characteristics \cite{Nelson2009}. 
Topography is an important factor affecting water distribution in soils since it directly affects overland flow and solar radiation rates \cite{Moore1991, Wilson2012, Tadono2014}. 
DSMs are the main inputs of geomorphometry, which is the science of quantitative land-surface analysis \cite{Pike2009}. 
The influence of topography on soil moisture prompts its present inclusion in soil moisture downscaling.

\begin{figure*}[hbp]
	\centering
	\frame{\includegraphics[trim={56mm 45mm 46mm 46mm},clip,width=\textwidth]{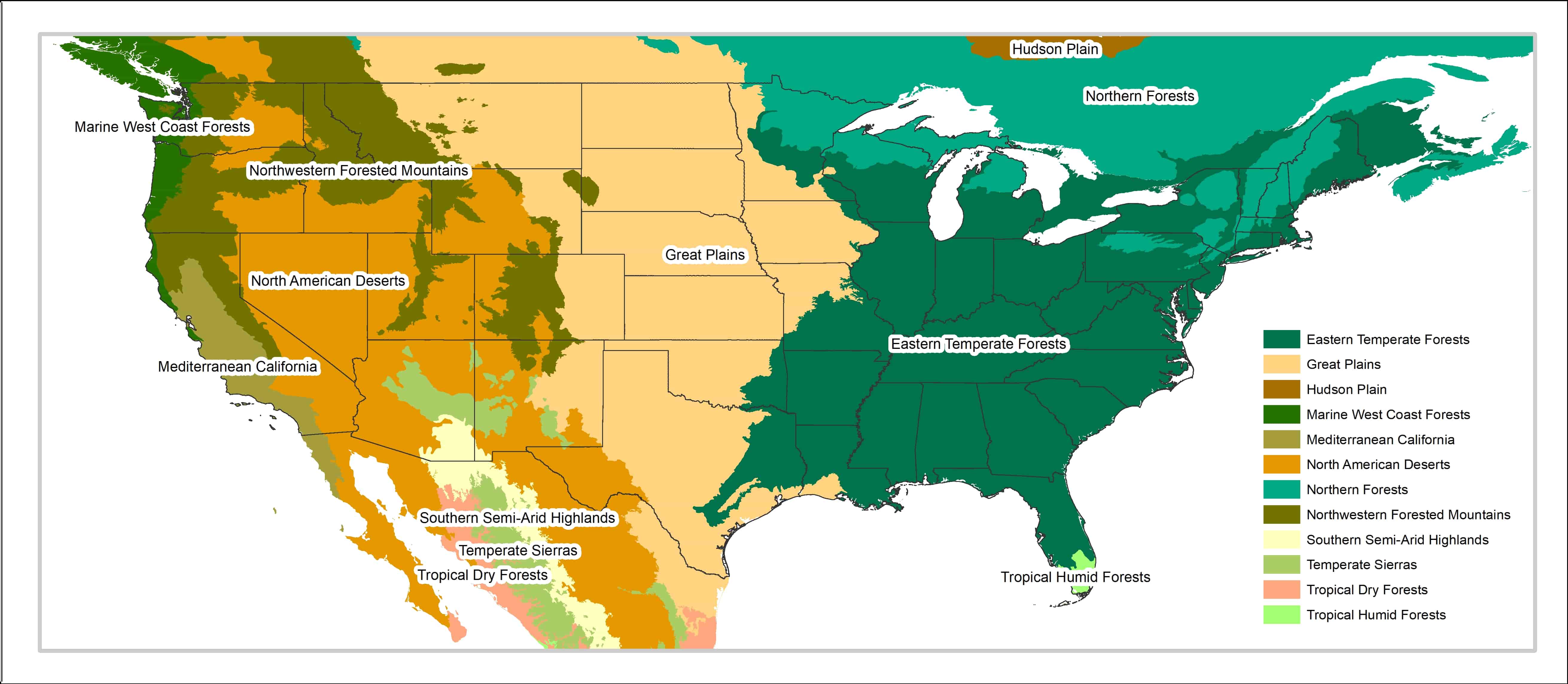}}
	\caption{CEC Level~I ecoregions across the conterminous United States.}
	\label{fig-lvl1}
\end{figure*}

To define the spatial limits of our soil moisture prediction, we use the 2011 update of the Commission for Environmental Cooperation (CEC) ecoregion dataset, developed jointly by Mexico, the United States, and Canada and based on the analysis of ecosystem elements such as geology, physiography, vegetation, climate, soils, land use, wildlife, and hydrology \cite{CEC2009}. 
This approach divides North America into polygon-based ecoregions at three levels, which range from Level~I (Fig.~\ref{fig-lvl1}), to Level~III and describe the similarity in the type, quality, and quantity of environmental parameters within the region. 
That is, Level~I regions are larger and more general and Level~III regions are smaller and more specific.

\section{SOMOSPIE Overview} \label{sec-model}

\begin{figure*}[htbp]
\centerline{\includegraphics[width=\textwidth]{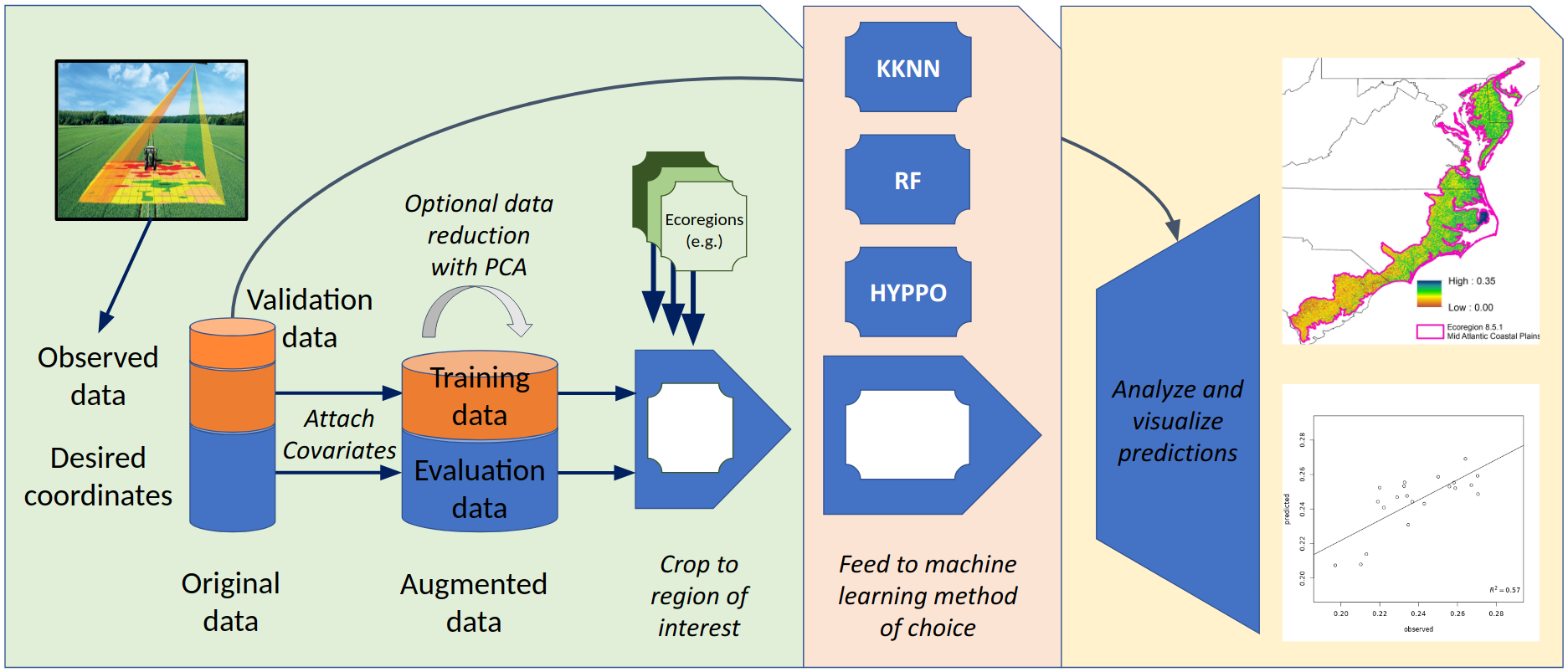}}
\caption{Overview of our modular SOil MOisture SPatial Inference Engine (SOMOSPIE) based on data driven decisions, including three prediction methods: kernel-weighted k-nearest neighbors (KKNN), Random Forests (RF), and Hybrid Piecewise Polynomial approach (HYPPO).}
\label{fig-workflow}
\end{figure*}

We build a modular SOil MOisture SPatial Inference Engine (SOMOSPIE) for prediction of missing soil moisture information. 
SOMOSPIE includes three main stages, illustrated in Fig.~\ref{fig-workflow}: 
(1) data processing to select a region of interest, incorporate predictive factors such as topographic parameters, and reduce data redundancy for these new factors; 
(2) soil moisture prediction with three different machine learning methods (i.e.,  KKNN, RF, and HYPPO); and 
(3) analysis and visualization of the prediction outputs.

\subsection{Data Processing} \label{A}

With SOMOSPIE, data are separated into two independent groups (i.e., observed data and evaluation data), to be fed into one of our modeling approaches defined in Section~\ref{B}. 
Observed data are represented as vectors, one for each pixel in the satellite data. 
Each vector consists of the latitude and longitude of the centroid of the pixel in the satellite data, an average soil moisture ratio for that pixel, and (optionally) the values of 15 topographic parameters from the digital surface model (DSM) evaluated at that centroid.
Additionally, a user can specify a percentage of the observed data to be randomly set aside as validation data. 
We discuss validation further in Section~\ref{C}. 
The remainder of the observed data not used for validation becomes our training data for generating models.
 
Evaluation data are represented as vectors, one for each pixel in the region of interest at a desired resolution.
Each vector consists of the latitude and longitude of the centroid of the pixel and (optionally) the values of the 15 topographic parameters from the DSM.
In this study, our desired resolution for soil moisture prediction is the $1 \times 1$ km resolution of the DSM we are using. 

Dimensional reduction of DSM data, when applied, is performed identically on both the training and evaluation data, as described below. 
The modeling techniques generate models using the training data and the models are evaluated on the evaluation data to generate the output prediction, as described in Section~\ref{B}. 

Topography and climate influence the spatial patterns of soil moisture \cite{Florinsky2012}. 
Our approach selects a specific ecoregion, with relatively similar environmental characteristics. 
Arguably, then, topography drives the predictions of soil moisture spatial patterns. 
Our engine uses the Commission for Environmental Cooperation (CEC) ecoregions as masks to select a specific region of interest and predict its soil moisture profile at a fine-grain resolution.

Our models use topography as the covariate space to downscale and gap-fill satellite-derived soil moisture, leveraging attributes derived from the DSM, such as the terrain slope or the aspect (i.e., the first and second derivatives of elevation data). 
These terrain parameters are surrogates of two main processes controlled by topography, the overland flow of water and the potential incoming solar radiation. 
Our topographic attributes are calculated using the SAGA GIS basic terrain parameters module (\url{saga-gis.org/saga_tool_doc/2.1.3/ta_compound_0.html}) \cite{Wilson2000, Wilson2012}. 

Because terrain attributes could have significant correlation, SOMOSPIE allows users to apply Principal Component Analysis (PCA) to reduce the number of covariates.  
This transformation is relevant because the reduction in DSM topographic parameters can reduce the time needed for prediction.  
To perform this reduction, the engine uses the PCA implementation from the Python package \emph{sklearn} (sklearn.decomposition.PCA), and selects components whose corresponding eigenvalues are at least one (a common rule-of-thumb~\cite{Wold1987}). 

\subsection{Prediction Models} \label{B}

SOMOSPIE presently supports three key machine learning modules for predictions of missing spatial soil moisture information: specifically kernel-weighted $k$ nearest-neighbor (KKNN), Random Forests (RF), and Hybrid Piecewise Polynomial (HYPPO). 
The selected methods have distinct character, with different  about the model being generated. 
Yet they all have one thing in common and that is built-in automated parameter tuning. 
KKNN tunes for the kernel and number of neighbors, RF tunes for number of variables per tree level, and HYPPO tunes for local polynomial degrees, all of which will be explained later.
Our implementations of all three methods use 10-fold cross validation to accomplish the tuning. 
This is a standard technique~\cite{Borra2010} which involves dividing data into 10 roughly equal parts. 
For every possible parameter value, 10 different models are generated, each using nine tenths of the data then being evaluated on the other tenth. 
The parameter value that minimizes cumulative error across all models is selected for generating a model with all the data. 
For our implementations of KKNN and RF, this tuning via cross-validation is performed with the R package \emph{caret}.
Having described the commonality of the methods, we now discuss the specific structure of each method in more detail.


The traditional $k$ nearest-neighbor ($k$NN) regression technique builds many simple models from local data.  
Use of this technique assumes that the $k$ points nearest in the prediction space to the data point one wishes to model are the most relevant and that points farther away have less influence on the point in question. 
The process begins with training data: points $\langle x^1_1,\ldots,x^d_1,z_1\rangle, \ldots , \langle x^1_n,\ldots,x^d_n,z_n \rangle$, where $(x^1_i,\ldots,x^d_i)$ are coordinates in the $d$-dimensional prediction space and $z_i$ is (in the present study) the corresponding soil moisture ratio. 
To predict soil moisture ratios for a specific choice of $(x^1,\ldots,x^d)$, $k$NN selects the $k$ nearest neighbors of $(x^1,\ldots,x^d)$ in the training data and uses the arithmetic mean of their associated soil moisture ratios.

A common generalization of $k$NN is to use a weighted mean of the $k$ nearest soil moisture ratios, where values from points nearer to $(x^1,\ldots,x^d)$ are given higher weights. 
The variant of $k$NN in our engine is kernel-weighted $k$ nearest-neighbors (KKNN)~\cite{Hechenbichler2004}, implemented with the R package \emph{kknn}. 
It uses a kernel function (i.e., rectangular, triangular, Epanechnikov, Gaussian, rank, or optimal) to compute neighbor weights for the mean. 
Cross-validation is employed (as described above) to determine the number of neighbors and which weighting kernel to use.

The Hybrid Piecewise Polynomial (HYPPO) module builds upon and extends traditional $k$NN in a different way to mitigating some of its limitations~\cite{Johnston2016}. 
Contrary to $k$NN, HYPPO allows local prediction models to be non-linear. 
In other words, the polynomial degrees in HYPPO become a flexible feature of the model. 
To build the prediction model with HYPPO, we start with $n$ training points, $\langle x_1,y_1,z_1 \rangle, \ldots,  \langle x_n,y_n,z_n \rangle$.  
We want to predict the value of $z$ (i.e., the soil moisture) for a new, specified coordinate $(x,y)$. 
Following the technique of $k$NN, we first find the $k$ nearest neighbors of $(x,y)$, then using the data of the $k$ nearest neighbors, HYPPO builds local models using a polynomial whose degree is selected using cross validation as described above. 
Na\"ive generation of a non-linear polynomial on many variables requires a large number of data points.
With 15 topographic parameters, the initial implementation of HYPPO would require a prohibitive number of neighbors. 
Therefore, as accommodation for more predictors is under development, the present study demonstrates HYPPO using only latitude and longitude as predictors.  

Random Forests (RF) consist of an ensemble of decision trees that are weighted via a statistical method called bagging. 
Each tree is grown with a random subset of predictors and of the training data. 
The tree's weight is determined by its 'out-of-bag error', which is computed by testing the tree on the rest of the training data. 
To make a prediction at a new point, all decision trees in the ensemble are queried and their predictions are combined using weighted arithmetic mean.
Such techniques do not assume a particular functional or geometric form of the model, and are thus suitable to deal with sparse datasets (e.g., areas with large gaps of soil moisture satellite estimates).  
SOMOSPIE employs the R package \emph{quantregForest} and has two main parameters: 
(1) the number of trees to grow in the ensemble of regression trees, and 
(2) the number of covariates randomly selected at each level of tree growth. 
In the present study, we consider a total of 500 trees for the first parameter. 
The second parameter is bounded above by the total number of prediction parameters--17 in our case with 2 spatial coordinates and 15 topographic predictors--and automatically selected using cross validation as described above.

\subsection{Analysis and Visualization} \label{C}

Our engine supports analysis nd visualization to assess model output.
As stated in Section~\ref{A}, a user can specify a percentage of the observed data to split off as validation data. 
All of the results in this paper were generated with 20\% of the observed datapoints randomly set aside as validation data. 
After generating predictions, the engine compares them to the validation points.
In this study, we use the squared correlation coefficient ($R^2$-value) as explained variance between the validation data at the coarse spatial resolution of $27 \times 27$ km (i.e., satellite-derived soil moisture) and the prediction of soil moisture at $1 \times 1$ km resolution. 
To accomplish this, we first harmonize the predicted data with the validation data by computing the arithmetic mean of all the predicted values for $1 \times 1$ km pixels that fall within a cell of the original grid ($27 \times 27$ km). 
To account for the randomness of the subset selected for validation, every execution of the workflow for this study is repeated 10 times.
For each execution, the resulting $R^2$-value is stored (rounded to two digits after the decimal); we give the arithmetic mean of the ten stored values. 
The standard deviation for the set of ten $R^2$-values fell between $0.088$ and $0.199$ in every case. 

The engine also provides a suite of visualization tools for soil moisture predictions and evaluations. 
It uses R and Python scripts to perform standard geographic information system (GIS) tasks (e.g., for both imagery and tabular forms of data). 
The current study demonstrates the creation of heatmaps for soil moisture predictions and scatter plots for comparisons of predictions to validation data. 
In both cases, rather than show all ten plots generated for every usage of the engine, we select an instance whose corresponding $R^2$-value is closest to the mean value for that usage. 
For the selected instance, we show the prediction heatmap as representative of the set of ten predictions.

\begin{figure}[htbp]
\centerline{\includegraphics[width=.5\textwidth]{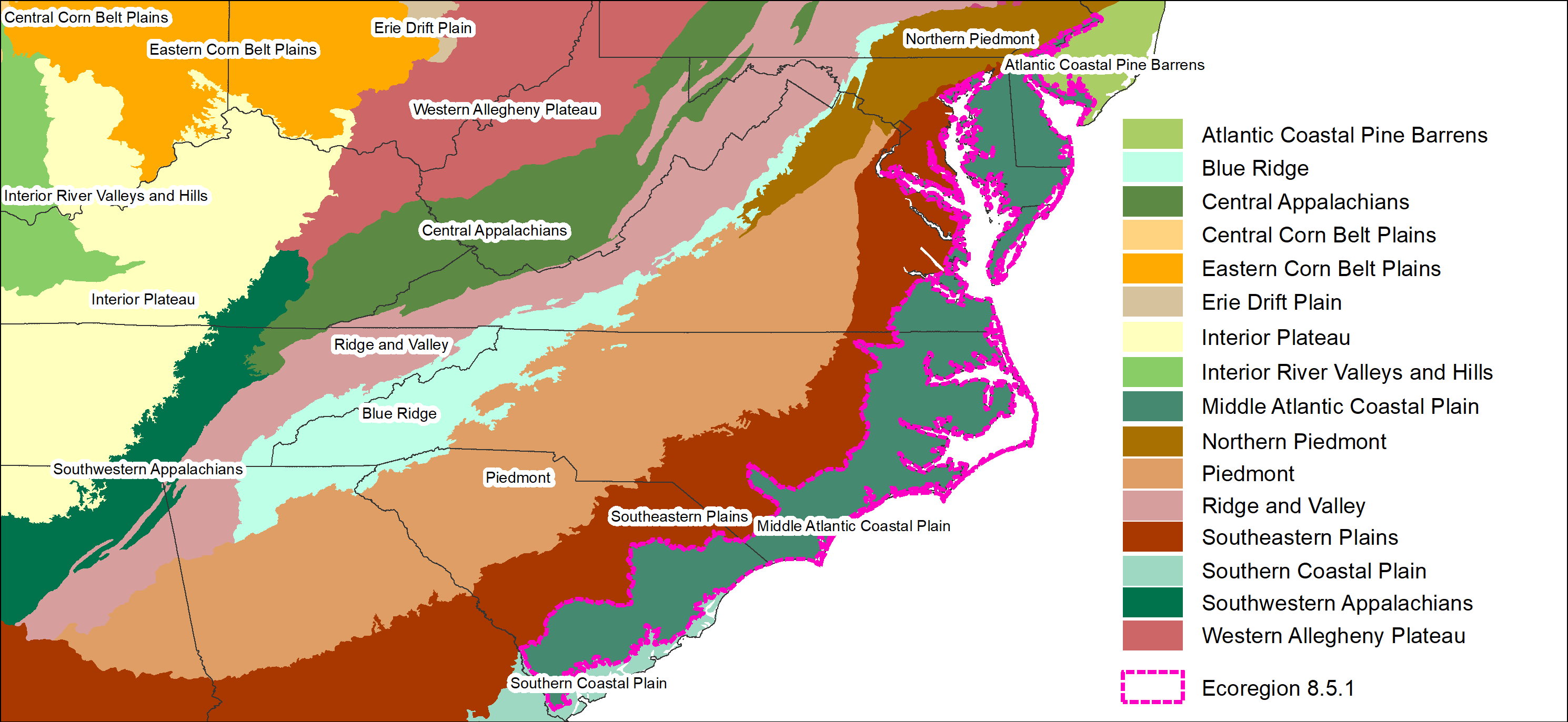}}
\caption{Selected Level~III ecoregion for this study: the Middle Atlantic Coastal Plains (8.5.1).}
\label{fig-lvl3}
\end{figure} 

\section{Prediction Results} \label{sec-results}

\begin{figure*}[hbp]
    \centering
    \subfloat[KKNN\label{fig-851-pred-kknn}]{\frame{
        \includegraphics[trim={1mm 1mm 0 1mm},clip,width=0.32\textwidth]{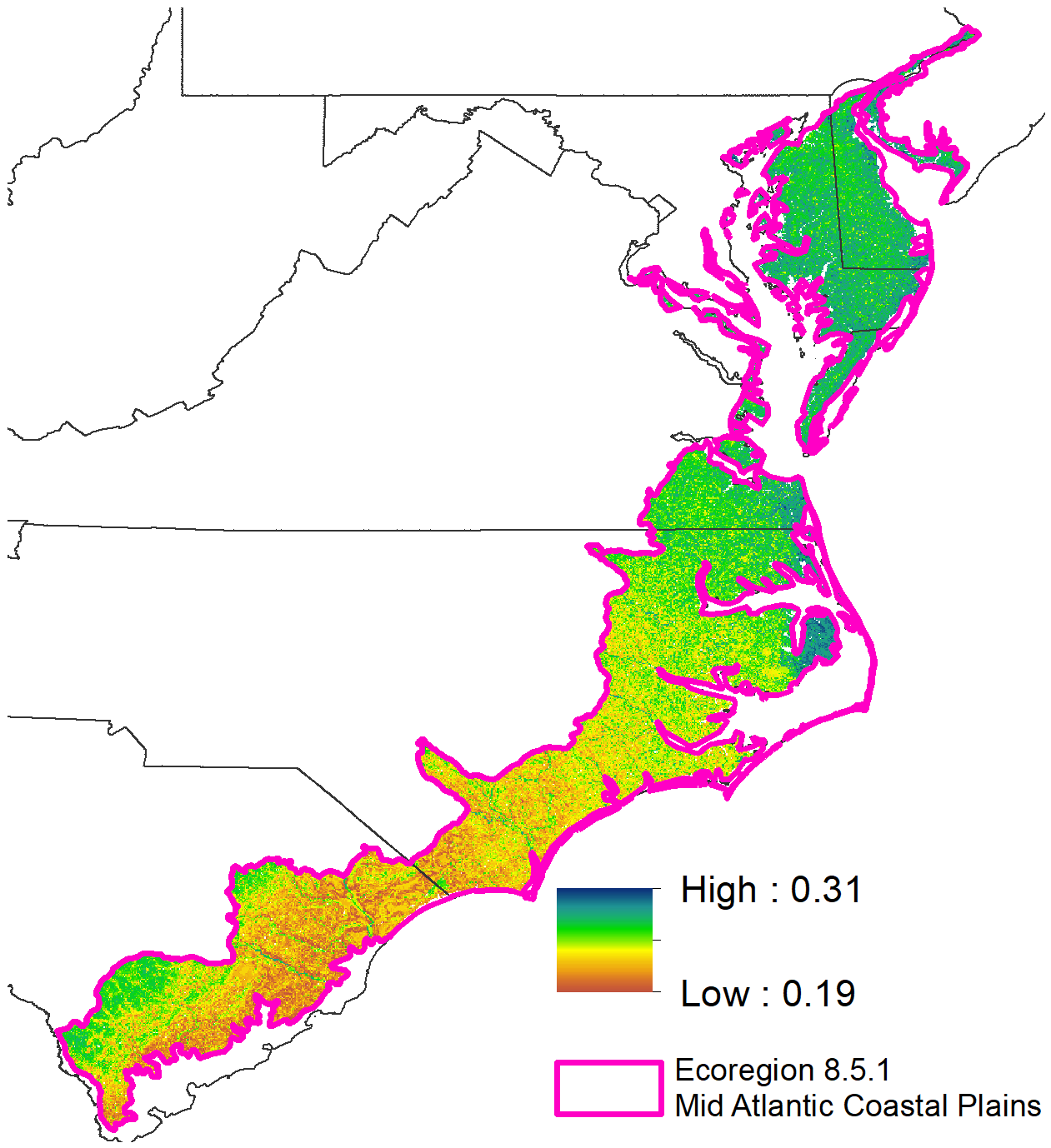}
    }}
    \subfloat[RF\label{fig-851-pred-rf}]{\frame{
        \includegraphics[trim={1mm 1mm 0 1mm},clip,width=0.32\textwidth]{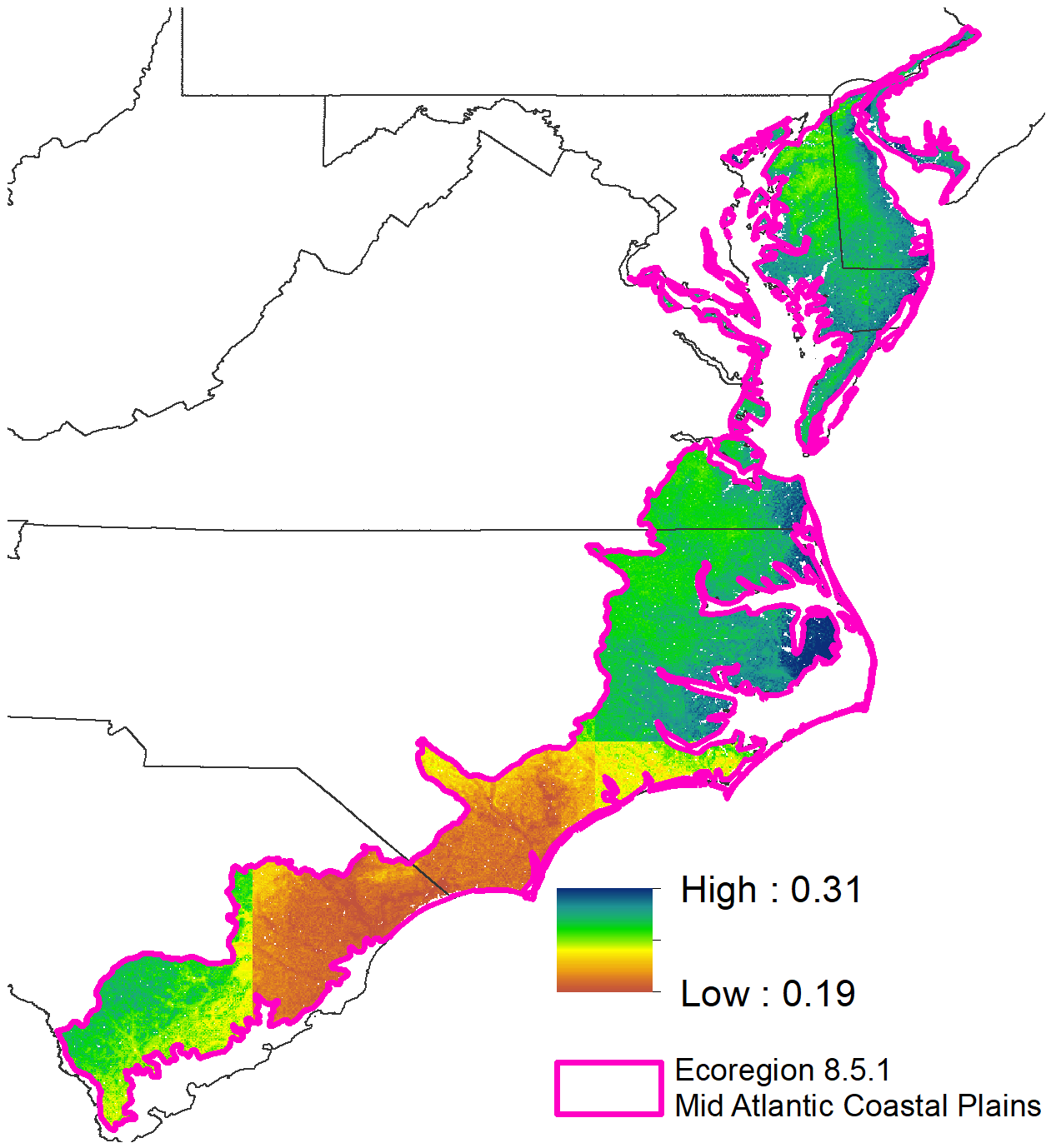}
    }}
    \subfloat[HYPPO\label{fig-851-pred-hyppo}]{\frame{
        \includegraphics[trim={1mm 1mm 0 1mm},clip,width=0.32\textwidth]{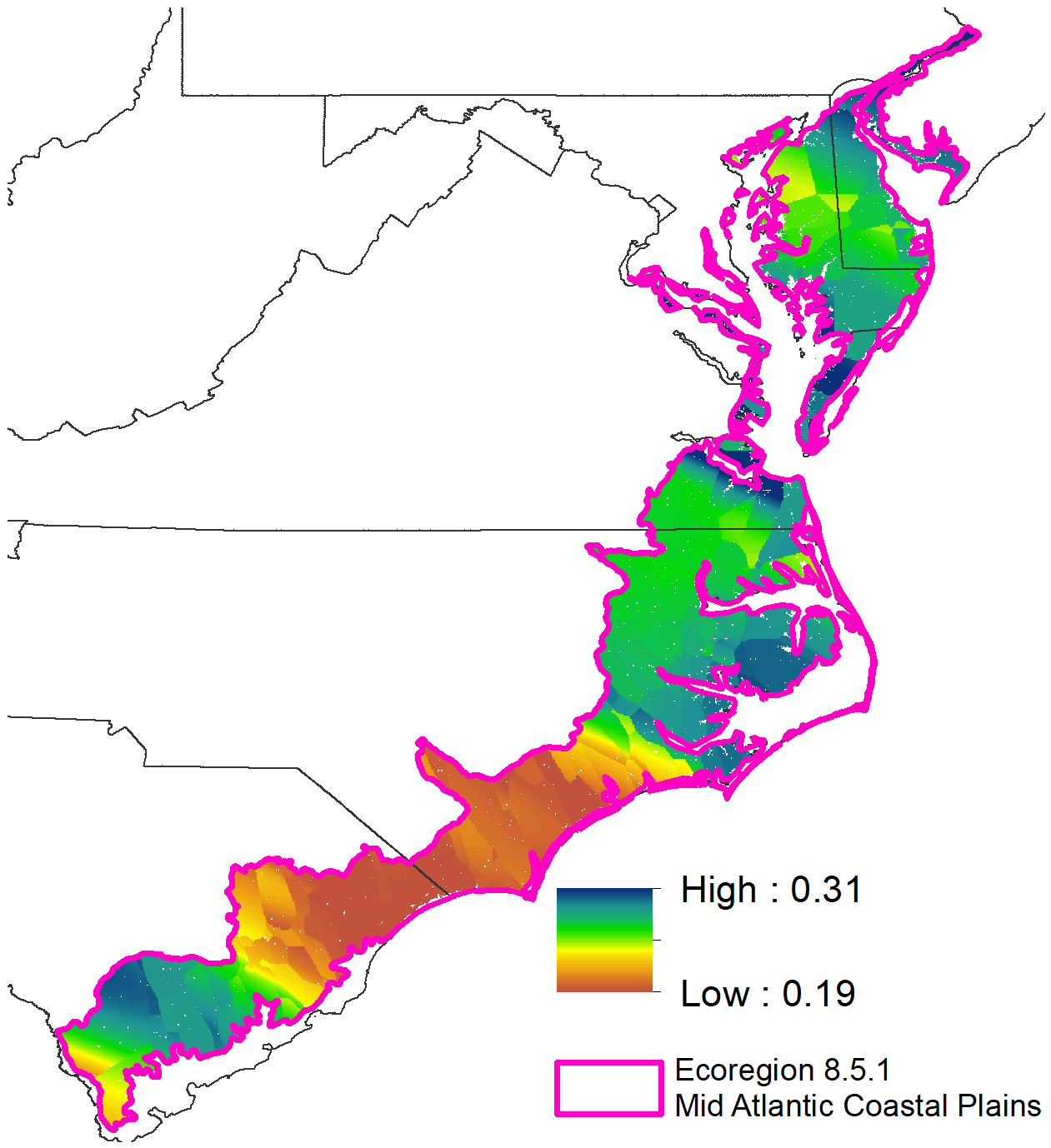}
    }}
    \caption{Example prediction maps ($1 \times 1$ km resolution) for ecoregion~8.5.1 (i.e., the Middle Atlantic Coastal Plains) generated by three different ML algorithms.}
    \label{fig-851-pred}
\end{figure*} 

For this study we selected a small Commission for Environmental Cooperation (CEC) ecoregion: the Middle Atlantic Coastal Plains (Level~III ecoregion~8.5.1, see Fig.~\ref{fig-lvl3}). 
This region has a broad range of moisture ratios with which to test the capabilities of SOMOSPIE.  
The soil moisture ratios we use for the observed data of all of these demonstrations are from April 2016. For each latitude and longitude coordinate pair in the satellite data, we take the average of all soil moisture ratios available in that pixel that month.

SOMOSPIE can increase the granularity of soil moisture information from coarse-grained satellite data to arbitrarily fine resolution using machine learning techniques. 
Presently we downscale from the original satellite-derived soil moisture native resolution ($27 \times 27$ km) to the $1 \times 1$ km resolution of our topographic predictors. 
In Section~\ref{model_decisions}, we demonstrate this with three machine learning methods: KKNN, RF, and HYPPO. 
Then in Section~\ref{data_decisions}, we take a step back in the SOMOSPIE workflow and examine the effects of data decisions on prediction.
In particular, we generate predictions for the same ecoregion (8.5.1) using KKNN and RF, but with training data from a larger region or with PCA reduction applied to the 15 topographic dimensions of the training data. 

\subsection{Soil Moisture Predictions} \label{model_decisions}

We now present soil moisture prediction on our local Level~III ecoregion (i.e., Middle Atlantic Coastal Plains) as heatmaps with soil moisture ratios between 0.19 and~0.31.
Fig.~\ref{fig-851-pred} shows predictions generated by our engine with three supported machine learning methods. 
Note the warmer colors in the southern portion (in South Carolina) representing lower ratios of moisture to land on the surface.

Overall, the pattern of soil moisture trends show an agreement across model predictions. 
We observe a more noisy prediction from KKNN, but unrealistic spatial artifacts (e.g., sharp jumps in North and South Carolina) from RF and, to a lesser extent, HYPPO. 
This is to be expected from HYPPO, since it is using only latitude and longitude as predictors and the coarseness of the original satellite data causes larger differences between observed soil moisture ratios of neighboring pixels. 
This indicates then that RF is more heavily effected by latitude and longitude than KKNN. 
We also observe the KKNN predictions having fewer extreme values (blue for wetter and dark orange for drier). 
So despite the spatial artifacts, RF and, to a greater extent, HYPPO, generally produce a diverse soil moisture trend that is more realistic for a region such as the Middle Atlantic Coastal Plains. 
This is consistent with the claim in \cite{Johnston2016} that HYPPO was ``designed to effectively and accurately model non-smooth [...] surfaces without the need for extensive sampling'' since ``most traditional techniques are designed to produce smooth models.''

\begin{figure*}[htbp]
    \centering
    \subfloat[KKNN\label{fig-851-corr-kknn}]{
        \includegraphics[trim={1mm 0 10mm 11mm},clip,width=0.348\textwidth]{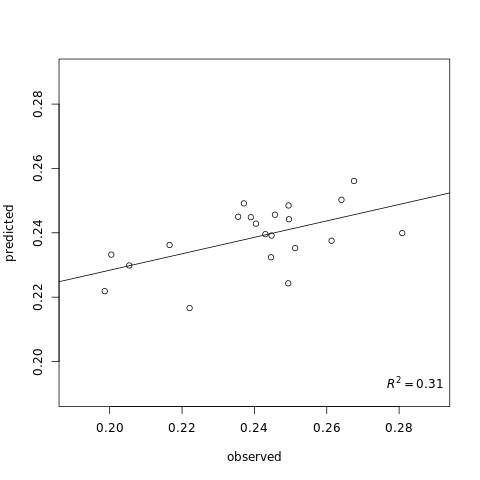}
    }
    \subfloat[RF\label{fig-851-corr-rf}]{
        \includegraphics[trim={20.6mm 0 10mm 11mm},clip,width=0.305\textwidth]{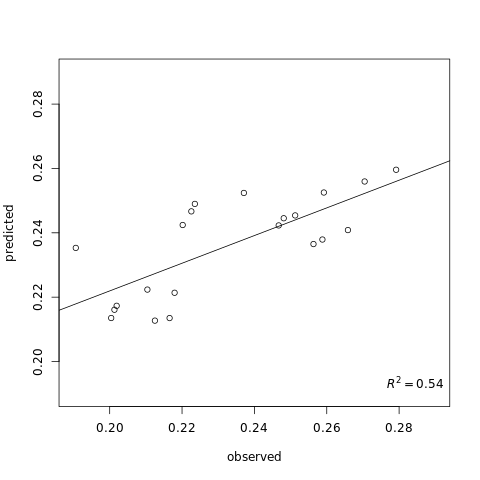}
    }
    \subfloat[HYPPO\label{fig-851-corr-hyppo}]{
        \includegraphics[trim={20.6mm 0 10mm 11mm},clip,width=0.305\textwidth]{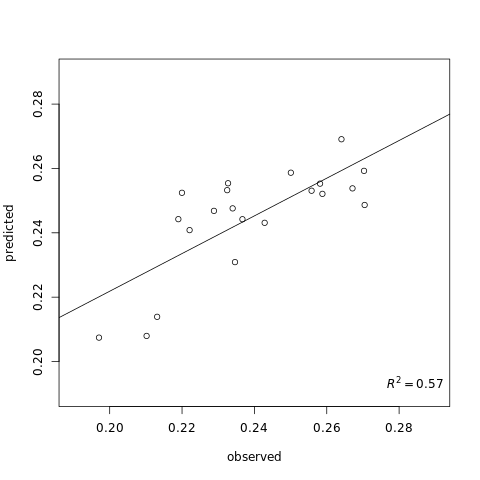}
    }
    \caption{Scatter plots comparing the modeled soil moisture from Fig.~\ref{fig-851-pred} and the validation data from the original satellite-based product ($27 \times 27$ km resolution).}
    \label{fig-scatter}
\end{figure*}

Moving on from qualitative observations, we investigate the quantitative relationship between the predicted soil moisture data and the initial coarse-grained observations (Fig.~\ref{fig-scatter}). 
We use explained variance ($R^2$-value) as an accuracy measure of the overall modeling performance, calculated from the relationship of the validation subset of the original satellite-based soil moisture data and the predicted soil moisture estimates. 
The KKNN models shows the lowest explained variance (a mean $R^2$-value of $0.296$) between observed and predicted satellite soil moisture ratios, and RF and HYPPO models have similar accuracy (with mean $R^2$-values of $0.575$ and $0.557$, respectively). 

\subsection{Impact of Data Processing Decisions} \label{data_decisions}

\begin{figure}[htbp]
\centerline{
    \frame{\includegraphics[height=100pt]{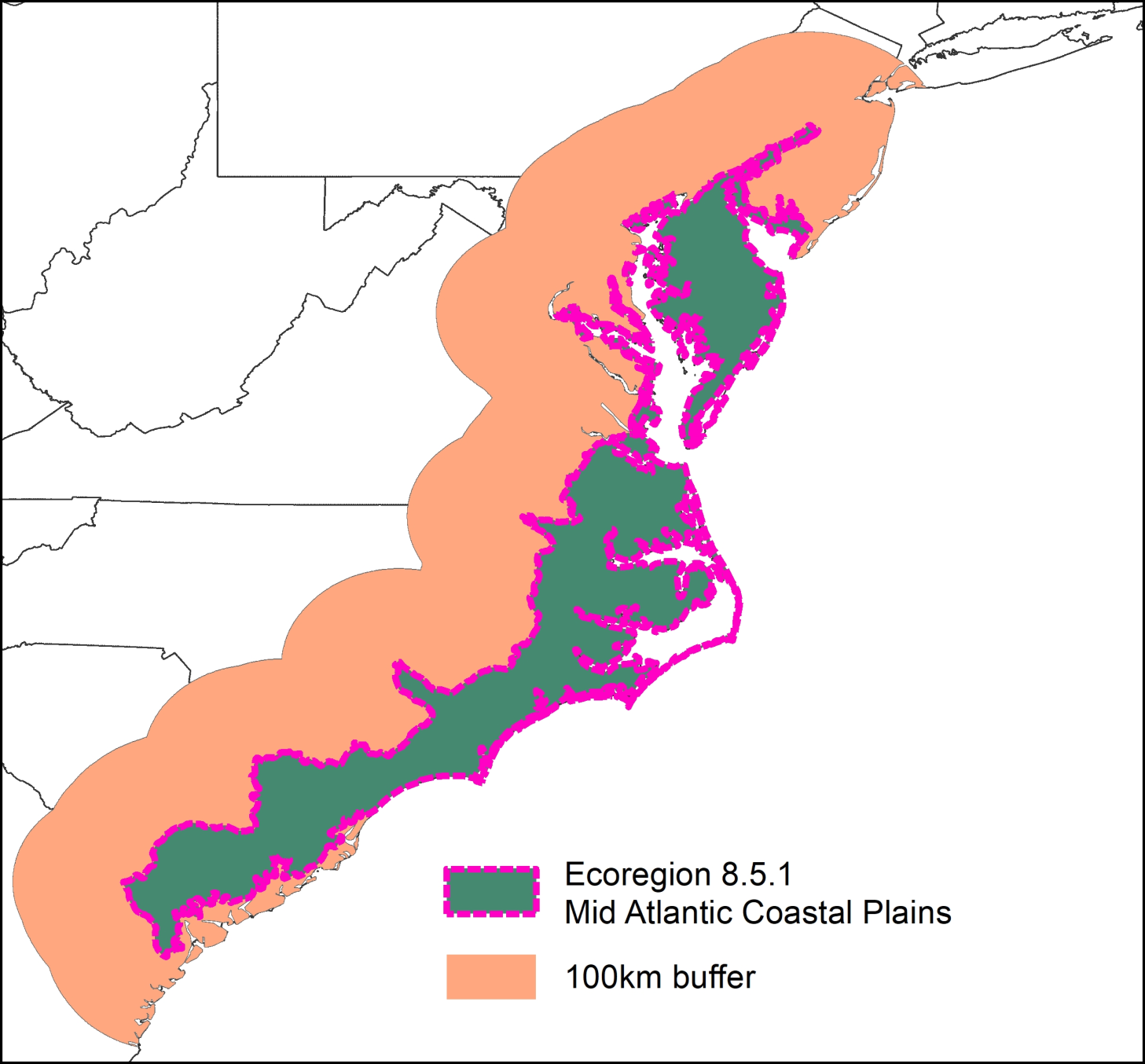}}
    \frame{\includegraphics[height=100pt]{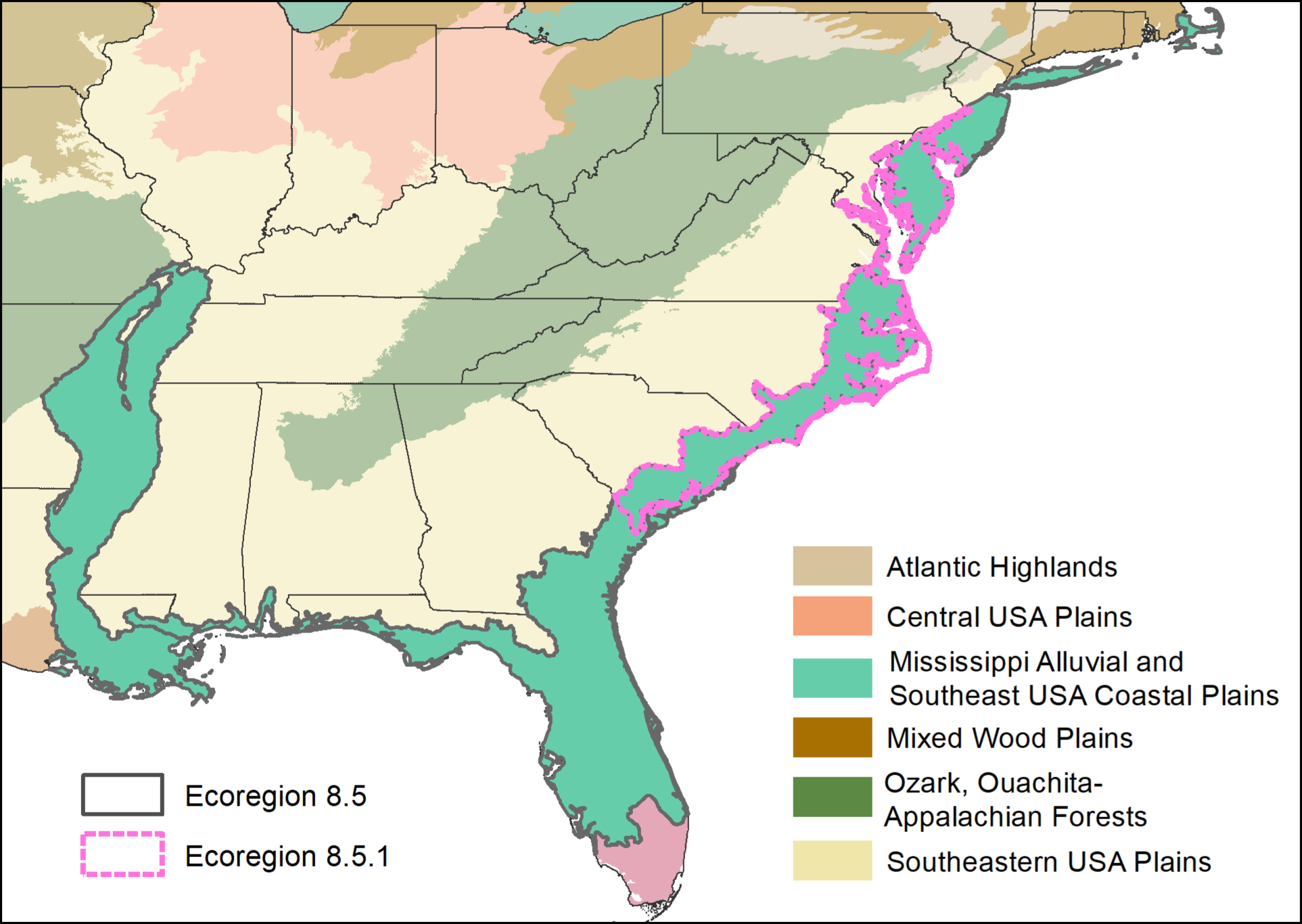}} 
}
\caption{Regions containing ecoregion~8.5.1, the Middle Atlantic Coastal Plains: 
    Level~III ecoregion~8.5.1 with a 100~km buffer (a, left); and
    Level~II ecoregion~8.5 (b, right).
}
\label{fig-super}
\end{figure}

One of the features in our engine when selecting a region of interest is the use of a larger region for model generation beyond the boundary of an ecoregion of interest.  
The assumption is that between neighboring ecoregions there is not necessarily a sharp separation but rather some sort of transition with multiple ecological gradients that serves as a buffer. 
Buffer selection could be useful for quantifying spatial gradients of ecosystem functional diversity and soil moisture feedbacks at the borders of ecological regions. 
Thus, the engine facilitates predictions using a larger region in two ways: 
(1) users can specify a buffer distance to be automatically added around the region of interest; or
(2) users can opt for prediction to use the lower level (larger, less ecologically specific) ecoregion containing the region of interest.
To investigate the effect of regional restrictions, we ran the same prediction methods (KKNN and RF) on three larger regions containing ecoregion~8.5.1, the Middle Atlantic Coastal Plains. 
Two of the three were ecoregion~8.5.1 with a fixed-width buffer, one with a 50~km buffer and one with a 100~km buffer (Fig.~\ref{fig-super}a). 
The third enlarged region is Level~II ecoregion~8.5, the Mississippi Alluvial and Southwest USA Coastal Plains (Fig.~\ref{fig-super}b).

\begin{figure*}[htbp]
    \centering
    \subfloat[Level~III 8.5.1\label{fig-super-pred-kknn-orig}]{\frame{
        \includegraphics[trim={1mm 1mm 0 1mm},clip,width=0.24\textwidth]{fig/KKNN_pred.png}
    }}
    \subfloat[8.5.1 with 50~km buffer\label{fig-super-pred-kknn-50}]{\frame{
        \includegraphics[trim={1mm 1mm 0 1mm},clip,width=0.24\textwidth]{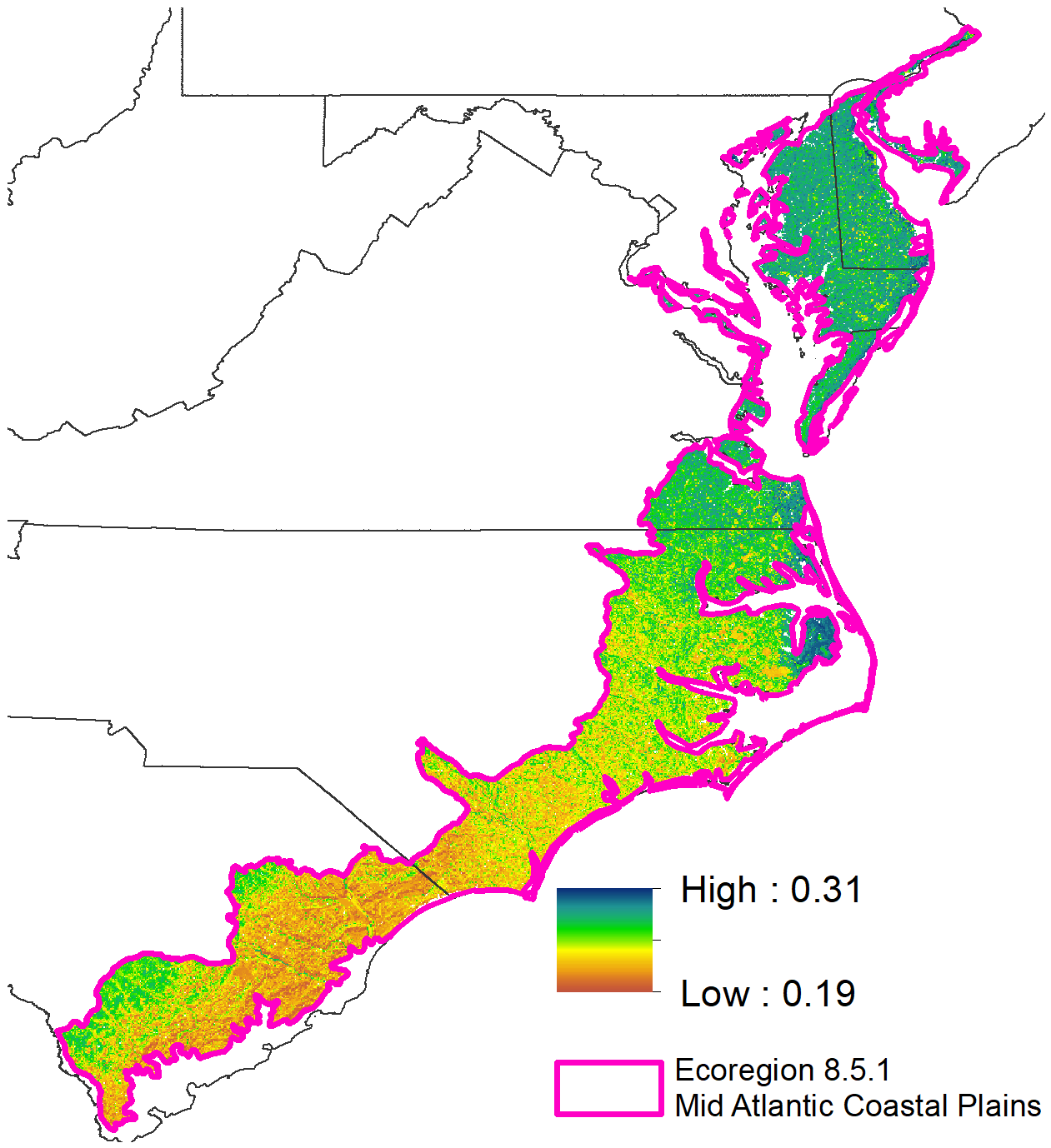}
    }}
    \subfloat[8.5.1 with 100~km buffer\label{fig-super-pred-kknn-100}]{\frame{
        \includegraphics[trim={1mm 1mm 0 1mm},clip,width=0.24\textwidth]{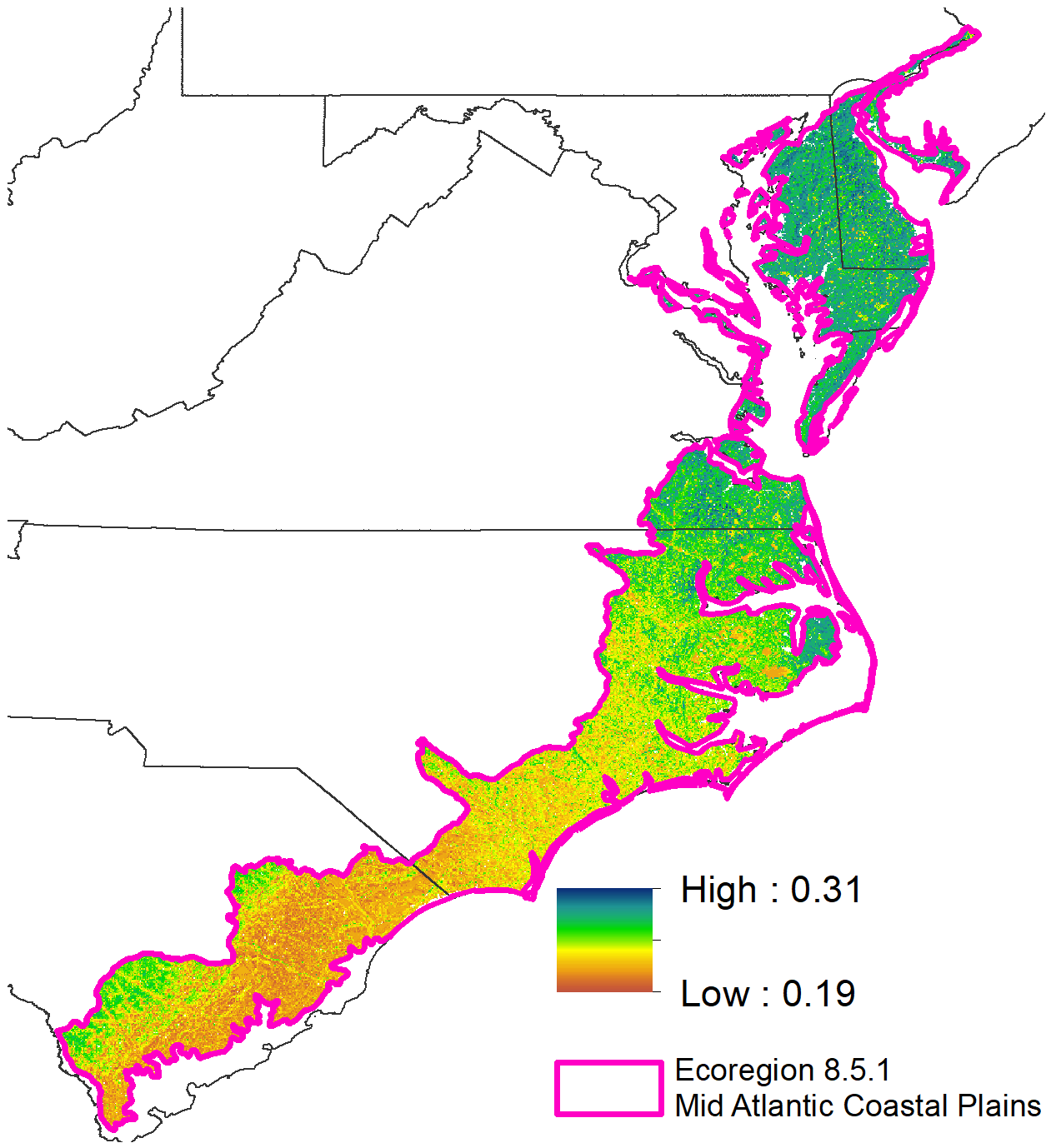}
    }}
    \subfloat[Level~II 8.5\label{fig-super-pred-kknn-lvl2}]{\frame{
        \includegraphics[trim={1mm 1mm 0 1mm},clip,width=0.24\textwidth]{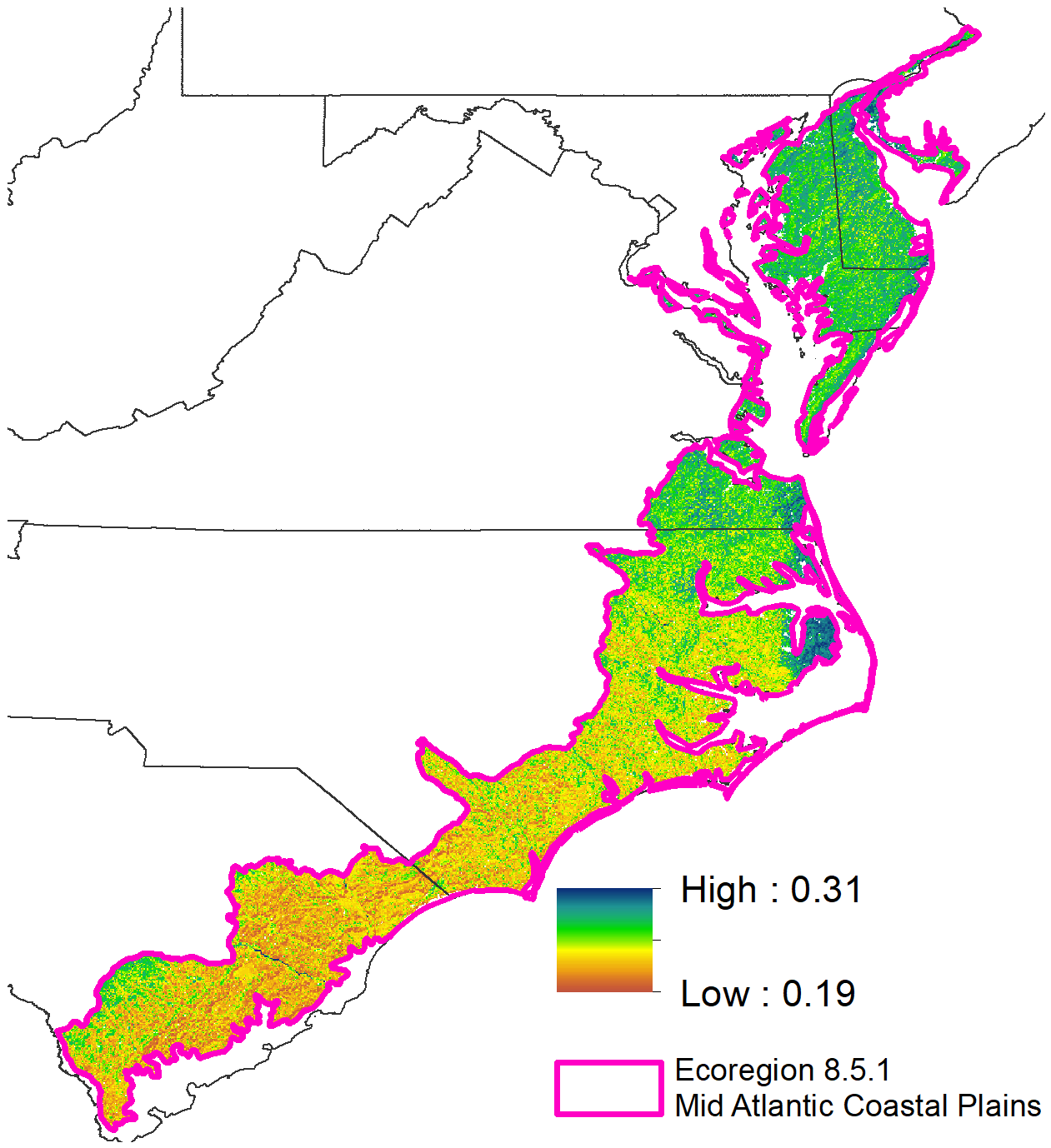}
    }}
    \caption{Prediction maps for the Middle Atlantic Coastal Plains (Level~III ecoregion~8.5.1) using KKNN on training data from areas containing Level~III ecoregion~8.5.1.}
    \label{fig-super-pred-kknn}
\end{figure*}

\begin{figure*}[htbp]
    \centering
    \subfloat[Level~III 8.5.1\label{fig-super-pred-rf-orig}]{\frame{
        \includegraphics[trim={1mm 1mm 0 1mm},clip,width=0.24\textwidth]{fig/RF_pred.png}
    }}
    \subfloat[8.5.1 with 50~km buffer\label{fig-super-pred-rf-50}]{\frame{
        \includegraphics[trim={1mm 1mm 0 1mm},clip,width=0.24\textwidth]{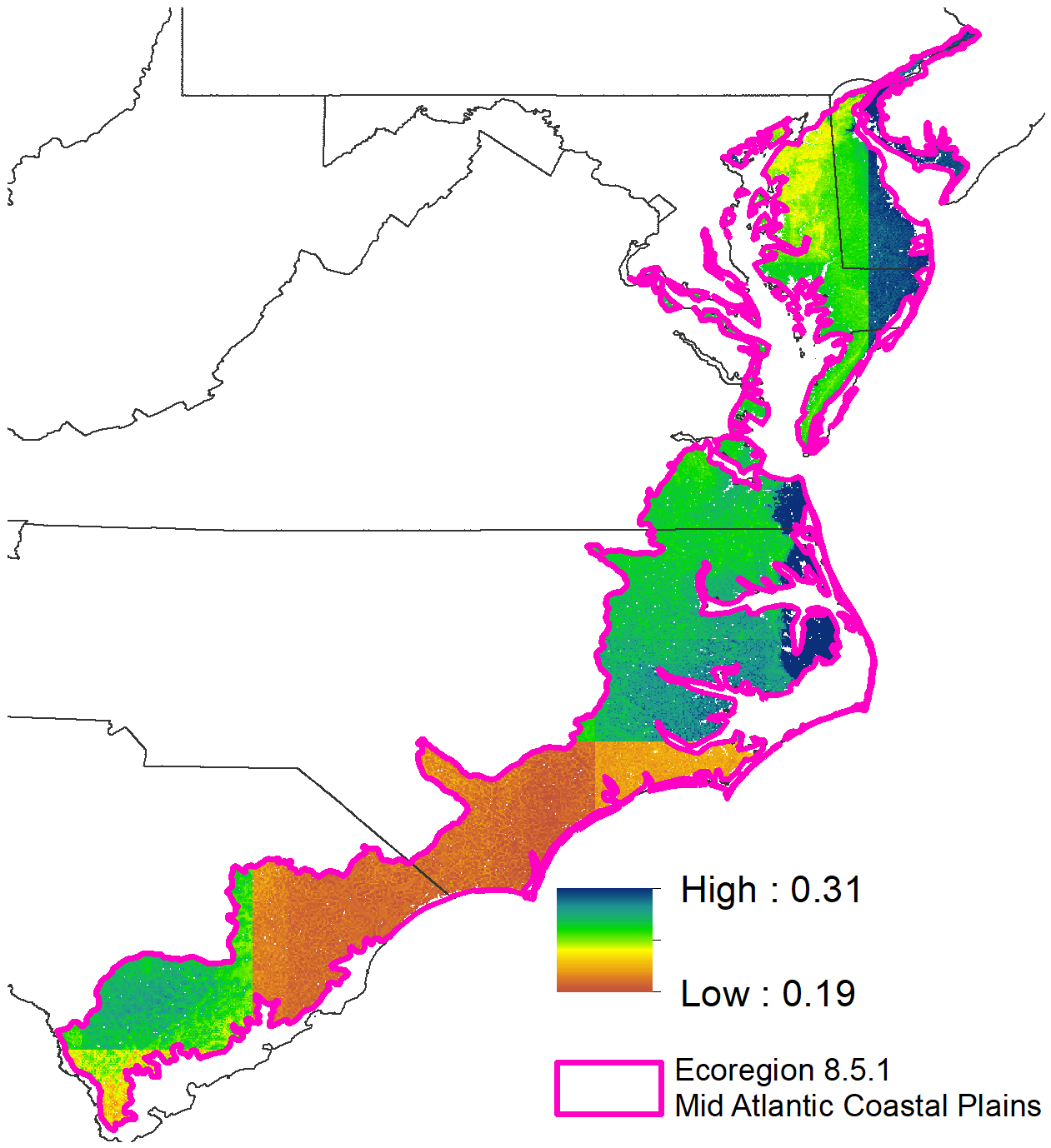}
    }}
    \subfloat[8.5.1 with 100~km buffer\label{fig-super-pred-rf-100}]{\frame{
        \includegraphics[trim={1mm 1mm 0 1mm},clip,width=0.24\textwidth]{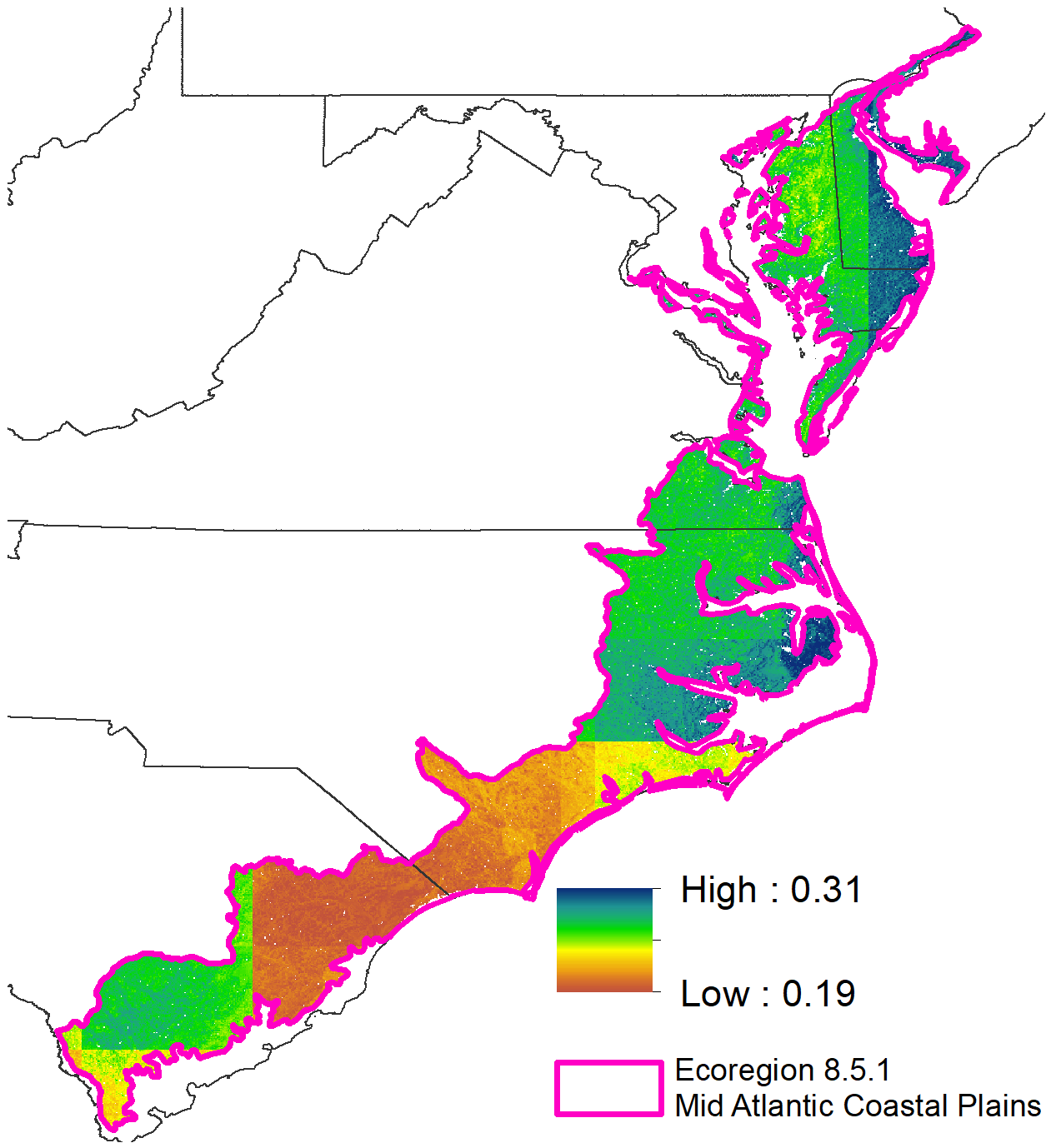}
    }}
    \subfloat[Level~II 8.5\label{fig-super-pred-rf-lvl2}]{\frame{
        \includegraphics[trim={1mm 1mm 0 1mm},clip,width=0.24\textwidth]{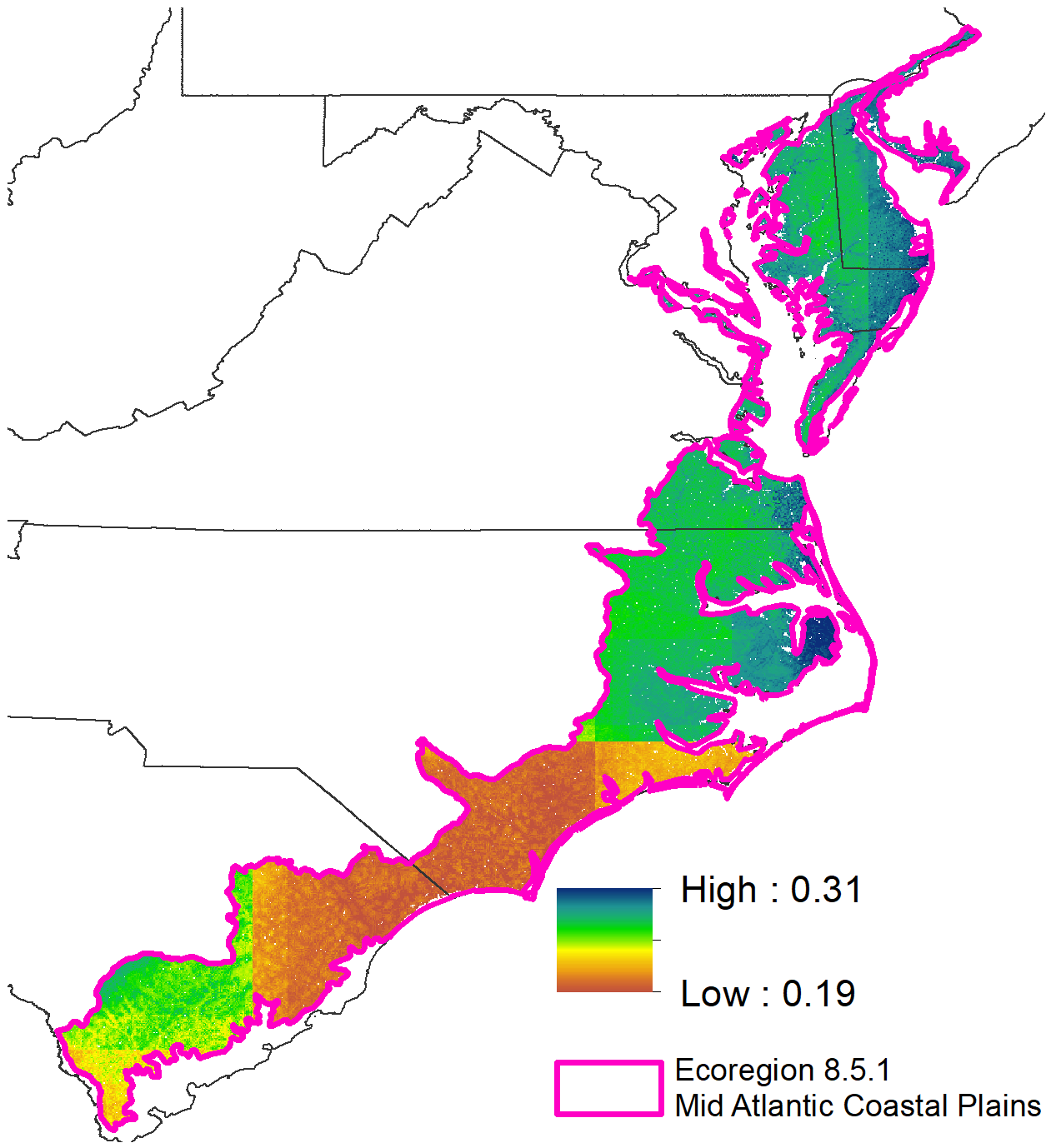}
    }}
    \caption{Prediction maps for the Middle Atlantic Coastal Plains (Level~III ecoregion~8.5.1) using RF on training data from areas containing Level~III ecoregion~8.5.1.}
    \label{fig-super-pred-rf}
\end{figure*}

The base results for comparison are from the predictions already generated for the region of interest, ecoregion~8.5.1, using KKNN (Fig.~\ref{fig-super-pred-kknn-orig}) and RF (Fig.~\ref{fig-super-pred-rf-orig}). 
We proceeded to generate models on the three larger regions, then evaluated the models to obtain soil moisture ratios only for our region of interest. 
When using a 50~km or 100~km buffer around the Level~III ecoregion, we observe slightly higher (darker green and blue-green) predictions in the northern third of the region (in Virginia, Maryland, and Delaware) in the KKNN models (Fig.~\ref{fig-super-pred-kknn-50} and~\ref{fig-super-pred-kknn-100}). 
Despite this shift, the spatial patterns are generally preserved across the KKNN predictions. 
On the other hand, when we add a 50~km or 100~km buffer, we see the spatial anomalies of RF models become more pronounced (Fig.~\ref{fig-super-pred-rf-50} and~\ref{fig-super-pred-rf-100}). 
When using training data from the entire Level~II ecoregion, KKNN appears to predict a narrower range of values in southern part of the region (Fig.~\ref{fig-super-pred-kknn-lvl2}) and RF still exhibits sharp lines but with smaller value jumps across those lines (Fig.~\ref{fig-super-pred-rf-lvl2}). 

For these models generated on various supersets of the region of interest, we now move from qualitative observations to quantitative analyses. 
We find that using training data further outside the region of interest sharply diminishes the explained variance between the KKNN predictions and the satellite observations. 
Whereas the original predictions had a mean $R^2$-value of $0.296$, that dropped to $0.240$ with a 50~km buffer, further to $0.204$ with a 100~km buffer, and even further to $0.099$ using the Level~II ecoregion~8.5.
However, the mean $R^2$-value for RF predictions was generally unaffected, with $0.575$ for the original region, $0.577$ with a 50~km buffer, $0.600$ with a 100~km buffer, and $0.551$ using the Level~II ecoregion. 
This indicates that KKNN is more prone to be negatively effected by extraneous data, yet supports the hypothesis that restriction to a region of common ecological character may be beneficial for some soil modeling efforts.

\begin{figure}[htbp]
    \centering
    \subfloat[KKNN\label{fig-pca-pred-kknn-before}]{\frame{
        \includegraphics[trim={1mm 1mm 0 1mm},clip,width=0.24\textwidth]{fig/KKNN_pred.png}
    }}
    \subfloat[KKNN with PCA\label{fig-pca-pred-kknn-after}]{\frame{
        \includegraphics[trim={1mm 1mm 0 1mm},clip,width=0.24\textwidth]{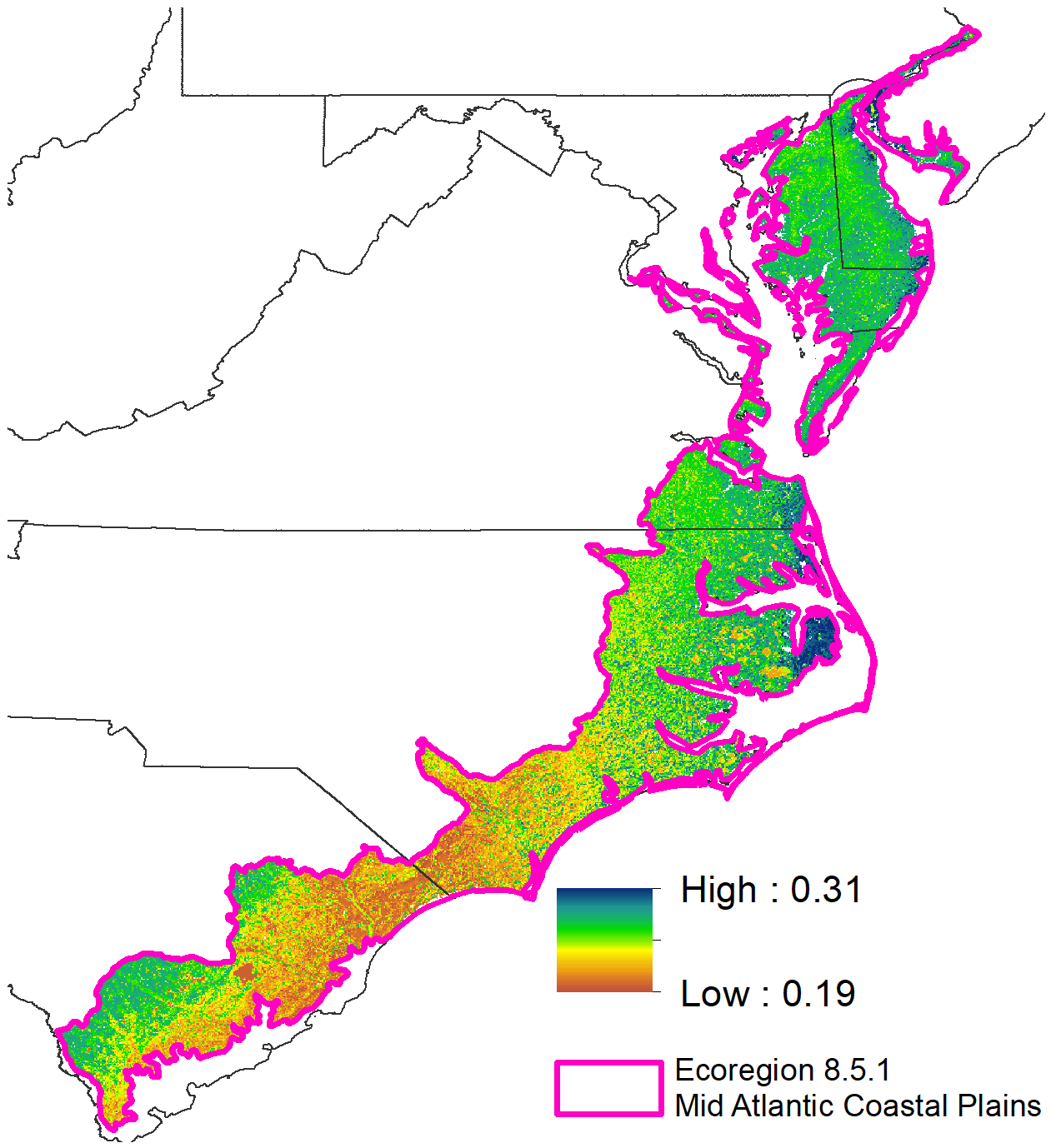}
    }}
    \\
    \subfloat[RF\label{fig-pca-pred-rf-before}]{\frame{
        \includegraphics[trim={1mm 1mm 0 1mm},clip,width=0.24\textwidth]{fig/RF_pred.png}
    }}
    \subfloat[RF with PCA\label{fig-pca-pred-rf-after}]{\frame{
        \includegraphics[trim={1mm 1mm 0 1mm},clip,width=0.24\textwidth]{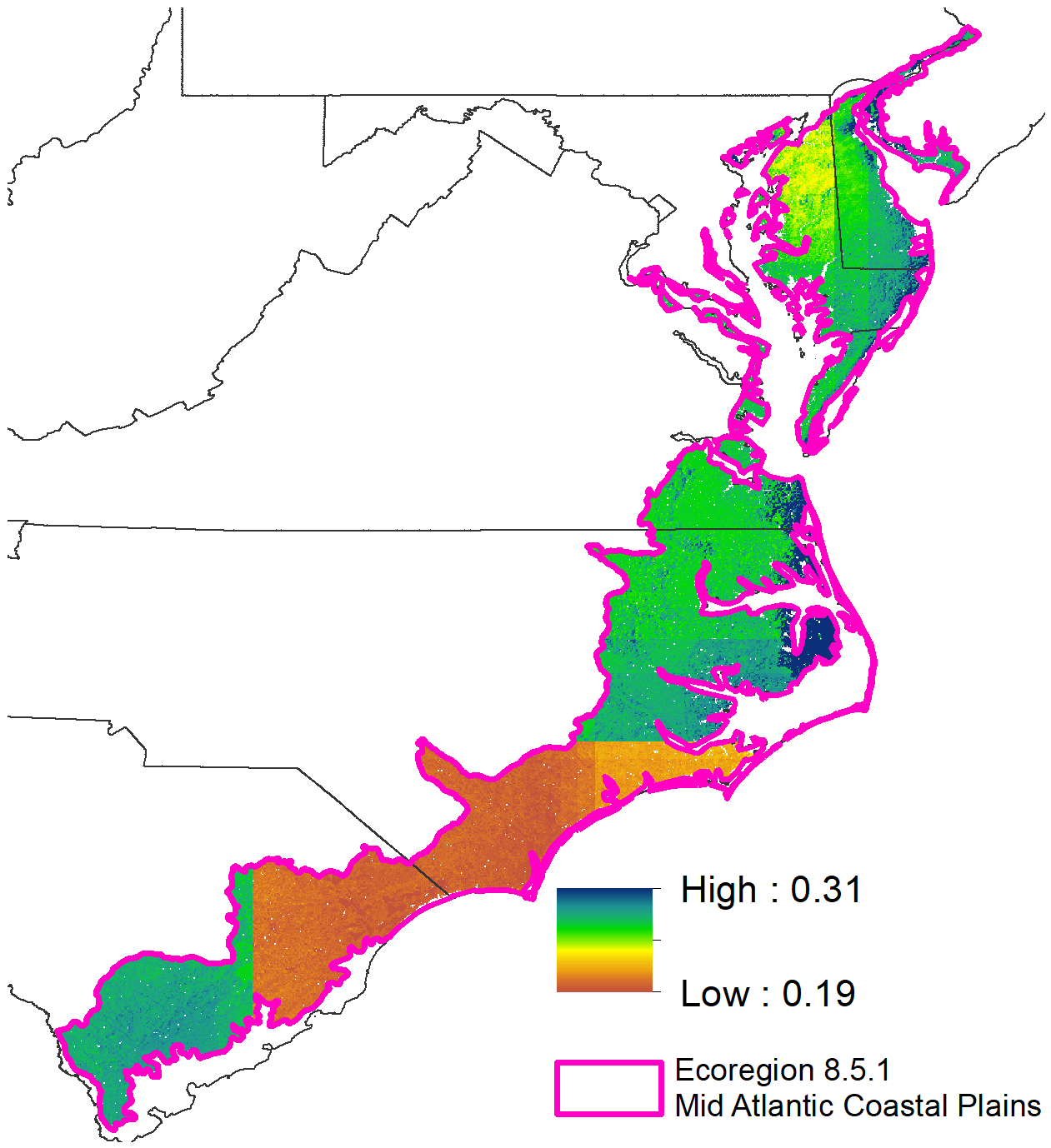}
    }}
    \caption{Effect of modeling over ecoregion~8.5.1 with KKNN (top) or RF (bottom) using all 15 DSM predictors (left) or with auto-selected PCA components (right).}
    \label{fig-pca-pred}
\end{figure}

Another optional features in our engine is the use of Principal Component Analysis (PCA) to reduce the number of covariates from terrain attributes before running a machine learning algorithm. 
We performed ten rounds of predictions with KKNN and RF over the region of interest on 80\% of the observed data with PCA reduction applied to the 15 topographic dimensions.
In nine of the rounds, 6 of the principal components were used (having eigenvalues above the standard threshold of one); 7 principals components were used in the other round. 
Fig.~\ref{fig-pca-pred} demonstrates the consequence of using the PCA dimension reduction on soil moisture prediction for the Middle Atlantic Coastal Plains. 
Specifically, Fig.~\ref{fig-pca-pred-kknn-before} and~\ref{fig-pca-pred-rf-before} show predictions when using all 15 DSM predictors while Fig.~\ref{fig-pca-pred-kknn-after} and~\ref{fig-pca-pred-rf-after} show the predictions with the PCA-reduced training data. 
In both cases, the use of PCA reduction appears to cause sharper local contrasts. 

To evaluate these predictions, we again compare the predicted soil moisture values with the validation subset of the observed satellite soil moisture estimates. 
We find that the KKNN model using all the topographic prediction factors showed only a slightly higher explained variance with a mean $R^2$-value of $0.296$ against the original product compared to the predictions based on the PCA transformation with a mean $R^2$-value of $0.289$.
However, we see a distinct rise in the mean explained variance for the RF predictions, which produced a mean $R^2$-value of $0.575$ with all the topographic dimensions, yet a mean $R^2$-value of $0.693$.
This indicates that a PCA dimension reduction could be useful to smooth out the data for modeling methods that tend to overfit.  
The PCA model significantly reduced the number of prediction factors (from 15 to 6 or 7), and thus reduced the computational demand for generating predictions across a $1 \times 1$ km grid. 
Both the statistical and computational performance are important considerations in data-driven modeling frameworks in order to maximize accuracy of results while constraining computation time.


Overall we observe how SOMOSPIE is able to perform different methods for predicting the spatial pattern of soil moisture across ecoregions with varying soil moisture profiles. 
The use and assessment of various modeling approaches is helpful to represent the complex variability of soil moisture and its dynamics, as each method is able to capture distinct elements of the soil moisture variability across the region of interest. 
Moreover, the study on the impact of data processing demonstrates how initial data processing decisions such as region selection can impact the prediction output, even for a fixed selection of machine learning model. 
Thus such decisions cannot be arbitrary and must be motivated by established scientific knowledge and by the accuracy of the predictions in comparison to existing data.

\section{Related Work} \label{sec-related}

Our work builds on recent technological advance in satellite-derived soil moisture (European Space Agency Climate Change Initiative~\cite{Wagner2012, Liu2011, Liu2012}). 
We reiterate that other satellite-derived datasets of soil moisture can be used in our workflow such as AMSR-E (Advanced Microwave Scanning Radiometer - Earth Observing System Sensor on the NASA Aqua Satellite~\cite{Owe2008}), ASCAT (Advanced SCATterometer aboard the EUMETSAT MetOp satellite~\cite{Wagner1999, Naeimi2009, Naeimi2012}), and AQUARIUS (Satellite instrument from NASA SMAP mission~\cite{Bindlish2015}). 
Despite technological advances, satellite datasets still have coarse spatial resolution and present temporal gaps making support tools such as SOMOSPIE useful to provide insights for research, environmental management, and precision agriculture based on remote sensing data.

Our work complements recent efforts~\cite{Pallipuram2014, McKinney2015} that provides the building blocks for interdisciplinary work and software development for soil moisture products. 
This project builds on the increasing recognition of the importance of spatial and temporal dependency of environmental data~\cite{Ettema2002, Dakos2011, Leon2014} and its application to precision agriculture~\cite{McBratney2005}. 
We take on the under-utilization of computer science techniques and computational resources to downscale satellite-derived soil moisture data, 
 in order to describe trends in soil moisture across CONUS. 
This project focuses on how topography and environmental variability influence soil moisture~\cite{Schmidhuber2007} across CONUS, and we postulate that our cyberinfrastructure tool has worldwide applicability to predict and downscale soil moisture from available coarse resolution satellite information.

\section{Conclusions} \label{sec-conclusion}

We developed the SOMOSPIE spatial inference engine for soil moisture data. 
This suite of cyberinfrastructure tools tackles the two main limitations of satellite-based soil moisture information: coarse granularity and spatial gaps. 
We demonstrate the potential of our engine by testing and comparing modeling decisions to predict the spatial variability of soil moisture across the Middle Atlantic Coastal Plains region of the United States. 
The modeling functionalities of SOMOSPIE include options for variable selection, data preprocessing, and method selection. 
Along with satellite-based soil moisture information, we integrate hydrologically meaningful prediction factors for soil moisture based on topography. 
Data preprocessing capabilities include training domain selection and data dimension reduction. 
For modeling method selection, our inference engine includes standard machine learning methods based on kernels (i.e., KKNN) and regression trees (i.e., Random Forests), and also integrates new modeling functionality with novel methods, such as HYPPO, not previously used for downscaling spatial data. 

To assess modeling decisions, SOMOSPIE includes tools for validation and visualization of output predictions. 
For our study region, KKNN performed poorly (with a mean $R^2$-value of 0.296 when predictions were compared to data from the original satellite observations) as compared to RF and HYPPO (with means $R^2$ values of 0.575 and 0.557, respectively).
We additionally demonstrate preprocessing decisions suitable for maximizing the effectiveness of data-driven soil moisture inference. 
For example, we use PCA dimension reduction on data fed to KKNN and RF; validation of the output indicates that PCA is a viable tool for lightening computational load without significantly affecting the result. 
On the other hand, expanding the area of training data beyond our ecologically specific region of interest negatively affected the prediction capabilities of KKNN; this indicates both the sensitivity of the particular method to the decision and the importance of carefully choosing one's training area within the prediction domain.  
Overall, we demonstrate the functionality of our tool to provide insights into where and why different methods yield distinct predictions.

Motivation for the SOMOSPIE system derives from the pressing need to improve spatial representation of soil moisture across the world for several applications in environmental sciences. 
Due to climate change (specifically, increasing temperatures) arid environments are expected to increase and ecosystem services (e.g., water and carbon cycling) across these areas may be at risk. 
Therefore, accurate soil moisture estimates are necessary to identify priority areas for soil resource conservation efforts and improve management decisions and Earth system models. 
Future work will consider validation against field observations (e.g., from the American Soil Moisture Database), comparison with other soil moisture information sources (e.g., NASA-SMAP), and more in-depth data-driven model tuning, with the ultimate goal to provide accurate soil moisture estimates using a globally applicable modeling framework.

\section*{Acknowledgment}

We note equal contributions of the first two authors, D.~R. and M.~G. 
We thank Travis Johnston for a Python implementation of HYPPO and Anita Schwartz for helping prepare the 15 topographic parameters from the DSM. 
We also thank Paula Olaya and Elizabeth Racca for contributing to preliminary prototyping, brainstorming, and experimenting during a summer 2017 research program at the University of Delaware.

\bibliographystyle{ieeetr}
\bibliography{refs}

\end{document}